\documentclass{article}

\PassOptionsToPackage{numbers, sort&compress}{natbib}

\usepackage[final]{neurips_2021}

\usepackage[utf8]{inputenc} 
\usepackage[T1]{fontenc}    
\usepackage{hyperref}       
\usepackage{url}            
\usepackage{booktabs}       
\usepackage{amsfonts}       
\usepackage{nicefrac}       
\usepackage{microtype}      
\usepackage{xcolor}         

\usepackage{mathrsfs}
\usepackage{mathtools} 

\usepackage{todonotes}

\usepackage{placeins}

\makeatletter
\renewcommand{\@noticestring}{%
1st Workshop on Human and Machine Decisions
(WHMD 2021) at NeurIPS 2021.}
\makeatother

\title{Catastrophe, Compounding \& Consistency in Choice}

\author{%
  Chris Gagne \\
  MPI for Biological Cybernetics\\
   \\
  Tübingen, Germany\\
  \texttt{christopher.gagne@tuebingen.mpg.de} \\
  \And
  Peter Dayan \\
  MPI for Biological Cybernetics\\
  University of Tübingen\\
  Tübingen, Germany\\
  \texttt{dayan@tue.mpg.de } \\
}
\begin{document}

\maketitle

\begin{abstract}

Conditional value-at-risk (CVaR) precisely characterizes the influence that rare, catastrophic events can exert over decisions. Such characterizations are important for both normal decision-making and for psychiatric conditions such as anxiety disorders -- especially for sequences of decisions that might ultimately lead to disaster. CVaR, like other well-founded risk measures, compounds in complex ways over such sequences -- and we recently formalized three structurally different forms in which risk either averages out or multiplies.  Unfortunately, existing cognitive tasks fail to discriminate these approaches well; here, we provide examples that highlight their unique characteristics, and make formal links to temporal discounting for the two of the approaches that are time consistent. These examples can ground future experiments with the broader aim of characterizing risk attitudes, especially for longer horizon problems and in psychopathological populations.

\end{abstract}

\subsection*{Introduction}

Given the many uncertainties in the present and future, we had to evolve sophisticated ways of handling risk. Individual appetites or aversion for risk differ substantially, with various forms of psychopathology arising at the extremes of these preferences.
Psychology and neuroscience have focused on single risky decisions (typically just one spin of the wheel of outrageous fortune). Historically,  heuristics dominated \citep{kahneman1979}; however, recently, axiomatically justifiable forms of risk sensitivity from the finance industry are starting to permeate. One such approach is known as conditional value-at-risk (CVaR$_\alpha$), which quantifies the risk of rare, extreme events or, more specifically, the expected value in lower $\alpha$-\% tail of the distribution \cite{artzner1999coherent}. Ecological risk, however, arises sequentially, over the long-run, as the decisions we make at one time ramify in the future. In this case, it is also important to consider subtleties in time consistency -- the most basic requirement for long-run planning that a decision-maker can ensure that their future self will voluntarily carry out their current intentions.

In recent work, we discussed three different long-run implementations of CVaR \citep{gagne_dayan_neurips}. Two of these, precommitted (pCVaR) and nested (nCVaR), lead to time-consistent choices and correspond respectively to evaluating risk always from the perspective of an initial, privileged state or re-evaluating future risk (whilst incorporating future evaluations) at every stage under the same risk level \citep{ruszczynski2006conditional,ruszczynski2010risk}. Fixed (fCVaR), which is not time consistent, arose in distributional reinforcement learning \citep{dabney2018implicit}, and involves looking at the same tail of the distribution at every stage.

We previously showed that taking account of risk using CVaR offers a better interpretation of the choices of a substantial body of human subjects in a simple, yet widely-investigated, sequential decision-making problem, the two-step task \citep{gillan2016characterizing,gagne_dayan_neurips}. We showed that genuine risk sensitivity can lead to apparently greater choice perseveration and lower estimates of learning rate, when risk is not modeled. This invites us to consider the role of risk more broadly in planning and sequential problems. To do so, however, we need to move beyond the two-step task, which was not designed either to determine risk preferences or to distinguish between (or reject) the various methods (precommitted, nested, fixed) of applying CVaR. In the current work, we simulate stylised examples allowing clearer discrimination and determination, showing:
(1) different risk-avoiders take structurally different trajectories around a potential catastrophe, with pCVaR, for instance, exhibiting the inverse of a `house money' effect; (2) nCVaR's blindness to the amount of bad luck experienced up to a current stage makes it adopt ever more negative risk evaluations given repeated, identical choices; and (3) in a simple problem with a constant hazard rate, nCVaR is equivalent to inflating this hazard rate inversely with the degree of risk aversion; and pCVaR to adopting a time-consistent sequence of hazard rates.

\subsection*{Conditional value-at-risk (CVaR)}

For a random variable $Z$, the conditional value-at-risk (CVaR$_\alpha$) is defined as the expected value in an $\alpha$-percent tail of a distribution. For the lower tail of a continuous distribution, it is defined as the average of the values lower than the $\alpha$-quantile ($q_\alpha$) (\cite{artzner1999coherent}):
\begin{equation}
\label{eq_def_CVaR}
\text{CVaR}_\alpha[Z] := E[Z | Z<q_\alpha(Z)] 
\end{equation}
$\alpha$ determines risk aversion by emphasizing the lower tail of the distribution more or less; for instance  $\text{CVaR}_{\alpha=0.1}$ corresponds to the expected value in the worst 10-\% of outcomes. $Z$ could be  a single random reward or cost, or the discounted sum of rewards minus costs (i.e., the return) in the case of sequential decision making.

When there are discrete outcomes, as in various of the problems we consider, the more general form of CVaR is used:
\begin{equation}
    \text{CVaR}_\alpha[Z] := \text{sup}_\nu \{\nu - \frac{1}{\alpha}E[(\nu - Z)^+]\}
\end{equation}

\subsection*{Precommitted, nested and fixed CVaR}

For sequential decision making, more specifically a finite-horizon MDP, the precommitted approach (i.e., pCVaR) is defined as:
\begin{equation}
\label{eq_def_pCVaR}
\text{pCVaR}^{\pi}_{\alpha_0,x_0} \coloneqq \text{CVaR}_{\alpha_0}[R_0 + \gamma R_1 + \gamma^2 R_2 +\dots |X_0 = x_0,\pi]
\end{equation}

with initial state $x_0$, rewards or costs $R_0, R_1, R_2, \dots$, and a policy $\pi$. Thus, the pre-committed approach corresponds to applying the `normal' CVaR to the entire sum of (potentially discounted) rewards, from the perspective of the initial state $x_0$. The risk under some policy $\pi$ is given by this pCVaR value. We write the risk level as $\alpha_0$, since recursive, Bellman-like, formul\ae\ for $\text{pCVaR}^{\pi}_{\alpha_0,x_0}$ rely on dynamic changes to $\alpha$ depending on the probabilities of the various outcomes at successive states \citep{chow2015risk}.

The nested approach (i.e., nCVaR), is defined as:
\begin{equation}
\label{eq_def_nested_CVaR}
    \text{nCVaR}^{\pi}_{\alpha,x_0} \coloneqq \text{CVaR}_{\alpha}[R_0 + \gamma \text{CVaR}_{\alpha}[R_1 + \gamma \text{CVaR}_{\alpha}[R_2 + \dots |X_2]|X_1]|X_0=x_0,\pi]
\end{equation}

This can be more easily interpreted inside out, rather like recursive calculation in Bellman evaluation; the CVaR at the risk preference ${\alpha}$ is applied to the last stage (here $R_2$ or beyond), and then the result of this evaluation is added to the reward in the previous stage; CVaR is applied again to this sum and this repeats until the initial state $x_0$. This nested approach has been theoretically motivated as a time-consistent conditional risk measure \citep{ruszczynski2006conditional,ruszczynski2010risk}.

For the fixed approach (i.e., fCVaR), CVaR is applied to the distribution of future rewards, like pCVaR in equation \ref{eq_def_pCVaR}. However, more like nCVaR, it is applied anew at each future decision stage, albeit to whole distributions of outcomes rather than the (random) scalars that are the nCVaR$_{\alpha}$ measures. While straightforward, this can lead to time-inconsistency, because of the way that the outcomes defining the tails of these distributions change across levels and decision stages.

The optimal policies for all three measures can be calculated using forms of dynamic programming, or more specifically backwards induction \citep{chow2015risk, ruszczynski2010risk, gagne}; see also Appendix B in \citep{gagne_dayan_neurips}. As noted, solving for the optimal pCVaR policy involves a dynamic adjustment of risk preferences over stages (which we then label as $\alpha_t$); in other words, pCVaR can look like the nested version (equation \ref{eq_def_nested_CVaR}), but with $\alpha$ replaced by $\alpha_t$, which differs potentially across time-steps and as a function of the random states and outcomes up to time $t$. This adjustment is made to keep track of the relevant portion of each state's distribution to the (precommitted) distribution of the start state; we discuss this adjustment as it arises in the specific examples.

\subsection*{(1) Navigation and an inverted `house money' effect}

We first consider differences between these three approaches in the grid-world shown in Fig.~1. Here, a decision maker starts in the bottom left hand corner and attempts to navigate to the goal on the right. Transitions are stochastic, so the lava pit (positioned along the bottom row) threatens direct paths. Risk aversion mandates more circuitous routes, to a greater degree for lower $\alpha$. This is true for all approaches (see Fig.~1; top row for fCVaR).

However, for a given $\alpha$, the three approaches make differing predictions (Fig.~1; bottom row). nCVaR is far more conservative, heading straight up to the top before going over and coming down. In contrast, pCVaR takes a riskier route in the beginning, but then stays further from the pit in the end. This is an inverted `house money' effect (the obverse of the gambler's fallacy) in which the felicitous avoidance of the pit early on leads, via a dynamic adjustment of risk preferences, to a decreased value of $\alpha_t$ later on. For the same level of $\alpha$, fCVaR is neither as conservative as nCVaR nor does it become more risk-averse at the tail end of the journey.

\begin{figure}
  \centering
  \includegraphics[width=1.0\linewidth]{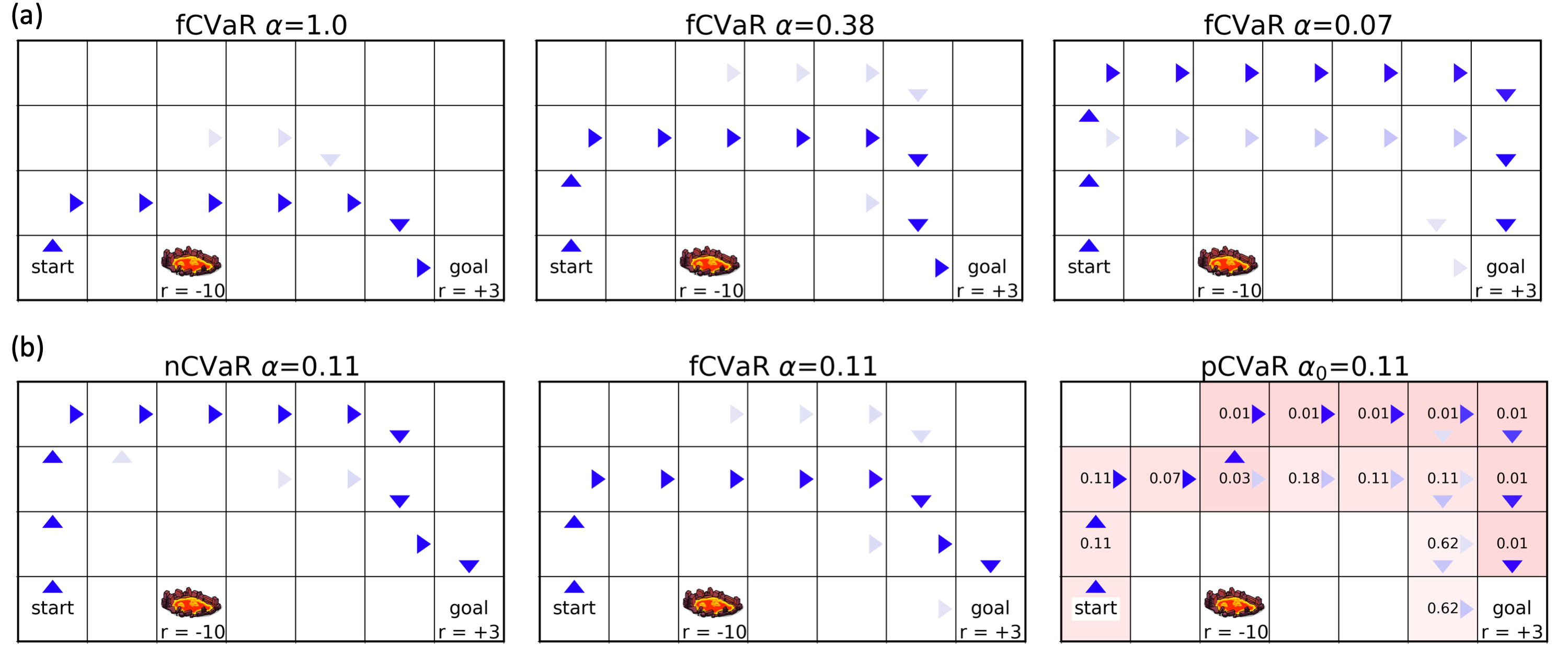}
  \caption{(a) fCVaR-policy for three values of $\alpha$; colored arrows represent frequencies (when $>5\%$) of actions taken in each square (only first visit to state shown). (b) CVaR-policy for n-, f- and pCVaR with the same $\alpha$. For pCVaR, the median adjusted $\alpha_t$ for each state is given by red shading.}
  \label{fig_1}
\end{figure}

\FloatBarrier

\subsection*{(2) The diverging effects of intermediate stages}

The navigation task focuses on how the optimal policy depends on the characteristics of risk aversion. How, though do the values of states change?  Fig.~2 (top) shows a simple Markov evaluation problem involving just four states, each with the possibility of some intermediate reward/loss (i.e., a 10-\% probability of getting a $-1$ and a 90-\% probability of getting a $+1$). We can expect that nesting, since it changes the structure of risk sensitivity, will lead to rather different consequences.

For fCVaR (and,  with stochastic fluctuations, pCVaR), all that matters is the future distribution of different possible combinations of rewards/costs. Even though, for $\alpha<0.2$, the contribution of the final stage can be negative, the fact that the intermediate stages are net favourable (-1 is less likely than +1) and the rewards in each stage are independent means that increasing the distance from the end increases the value (Fig.~2b). Intuitively, with more intermediate stages, the chances of rewards build while the chances of getting sufficient consecutive losses to overcome these rewards diminish.

For nCVaR, however, compounding works quite differently, and varies dramatically with $\alpha$. With modest risk aversion (high $\alpha$), increasing the number of stages increases the start value just like p/fCVaR. However, with more extreme risk aversion (low $\alpha$), increasing the number of stages actually decreases the initial value (Fig.~2c). If $\alpha<0.2$, the final stage $s=0$ is net negative; this quantity is then subtracted from all the possible outcomes at the state $s=1$, and so forth. Thus, the negative value will accumulate linearly going back to the start state. 

From a psychological perspective, this suggests that some  extreme risk preferences could arise from viewing the decision problem as a longer series of intermediate steps and evaluating the risk (unfavourably) in each. If so, this would predict that encouraging a broader perspective, i.e. a single distribution at the start (i.e.,  pCVaR), would induce a more mild risk preference. 

\begin{figure}
  \centering
  \includegraphics[width=1.0\linewidth]{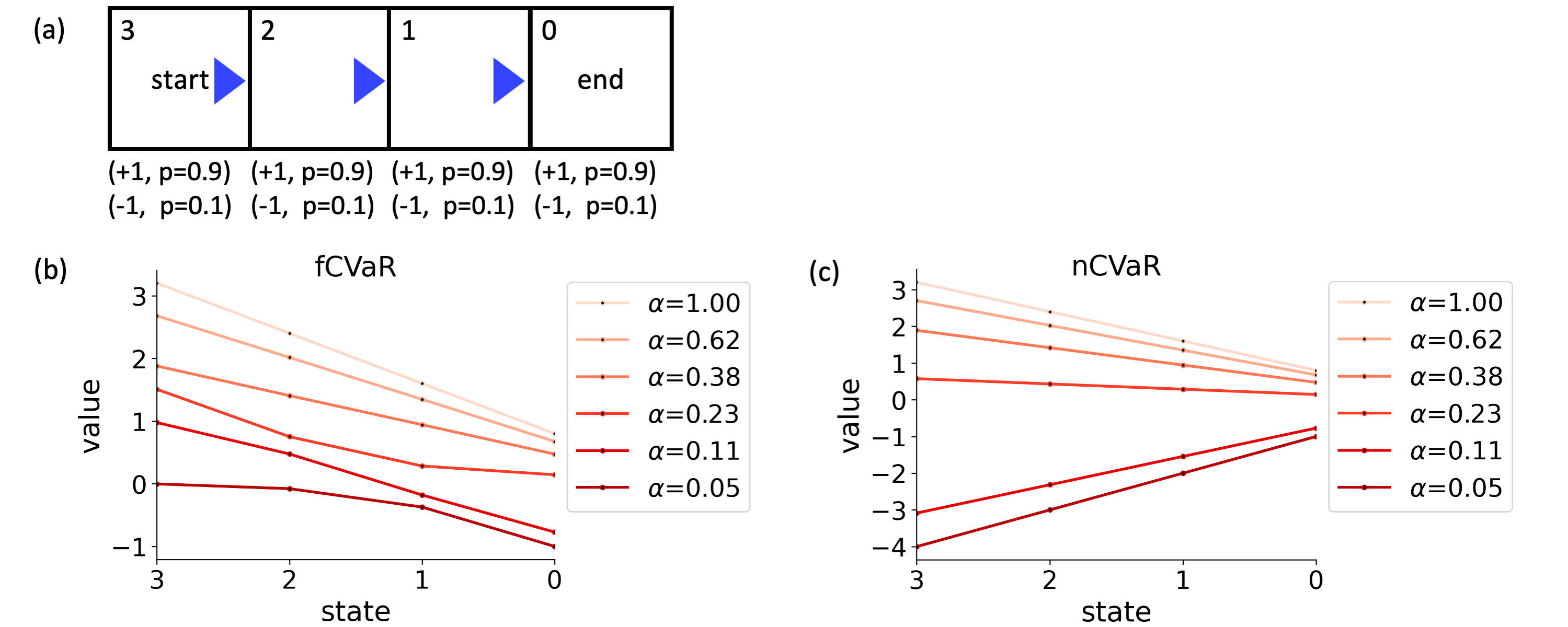}
  \caption{(a) A simple Markov evaluation problem with four states, each associated with possible reward/loss of +1/-1 with probabilities 0.9 and 0.1. (b) For fCVaR, the value increases with further distance from the end due to the net favorability of each intermediate state; this holds across $\alpha$-levels. (c) For nCVaR, the initial value either increases or decreases depending on the level of $\alpha$.}
  \label{fig_2}
\end{figure}


\subsection*{(3) Risk as discounting}

In the previous problem, values accumulated linearly across successive states. A final window into these three forms of risk sensitivities comes from considering the sort of multiplicative compounding that arises when a single reward (in Fig.~3a, $r=1$) 
is available from a sequence of lucky transitions. It is well known that geometric discounting (with value $\gamma$, say) in the standard, expected value case ($\alpha=1$) can arise from a fixed probability per step of terminating (a fixed hazard of $1-\gamma$). Given only rewards, increasing risk sensitivity (i.e., lower $\alpha$) implies steeper discounting, because the chance of dying looms larger on every step. However, how does this differ for the three approaches? The answer turns out to shed additional light on the time consistency of n- and pCVaR.

We can first compare the approaches by plotting their values for each state and comparing with conventional geometric discounting for different values of $\gamma$ (Fig.~3b-d). For $\alpha=1$, all the methods match geometric discounting. For fCVaR, for greater risk sensitivity (e.g., $\alpha=0.11$),  discounting starts more steeply than geometric discounting but then becomes shallower (Fig.~3b); in other words, it cut across different values of $\gamma$ in rather the same manner as hyperbolic discounting. For pCVaR, the reverse pattern holds for low $\alpha$ (Fig.~3c). Strikingly, however, for nCVaR, discounting remains perfectly geometric for any $\alpha$, albeit with a different, lower, value of $\gamma'$ (Fig.~3d). For example $\alpha=0.1$ matches with $\gamma'=0.5$.

This arises in the particular problem because of multiplicative compounding. Consider the penultimate state 1. For $\alpha=0.1$, the $5\%$ chance of dying takes up half the lower tail, and so is inflated to be $0.5$. Thus the value is $0.5\times [r=1]=0.5$. From state 2, this value is then again multiplied by the deflated chance of surviving to $0.5\times 0.5=0.25$; and so forth. For all $\alpha>(1-\gamma)$, the relationship between the implied discount factor and the risk-preference (and the hazard rate) is: $ \gamma' = (1-\nicefrac{(1-\gamma)}{\alpha})$. Thus, the risk sensitivity for the nested approach acts to increase the apparent hazard rate in exact proportion to the tail determined by $\alpha$. This also gives some more intuition to why the nested approach is time-consistent (discussed in more detail in \citep{gagne_dayan_neurips}); it is well known that geometric (or exponential) discounting leads to time-consistent preferences, while alternatives such as hyperbolic discounting do not. Here, the nesting allows the risk to be propagated back in just such a fashion. In contrast, the fixed approach is not time-consistent, as is discussed elsewhere; e.g. in \citep{gagne_dayan_neurips}. 

Revisiting pCVaR, the values at each state did not correspond to $\gamma^t$ for any fixed $\gamma$. However, the recursive logic of pCVaR in this simple problem means that each state $t$ has an effective discount value of $\gamma'_t=1-\nicefrac{(1-\gamma)}{\alpha_t}$, where $\alpha_t$ is the adjusted value required by pCVaR to keep its precommittment. For example starting at $\alpha=0.3$, we have $\alpha_3=0.3, \alpha_2=0.26, \alpha_1=0.22$ \footnote{As also seen in the navigation problem, where good fortune (i.e., survival) is followed by increased risk-sensitivity.} and, in effect, the discounting is less extreme at the start ($\gamma'_3=0.83; \gamma'_2=0.80; \gamma'_1=0.77$). This sort of progressive discounting is also time-consistent (\citep[][Thm 13]{lattimore2014general}, because at each time step, it uses a single series of future discount factors that can be planned from the beginning.

\begin{figure}
  \centering
  \includegraphics[width=1.0\linewidth]{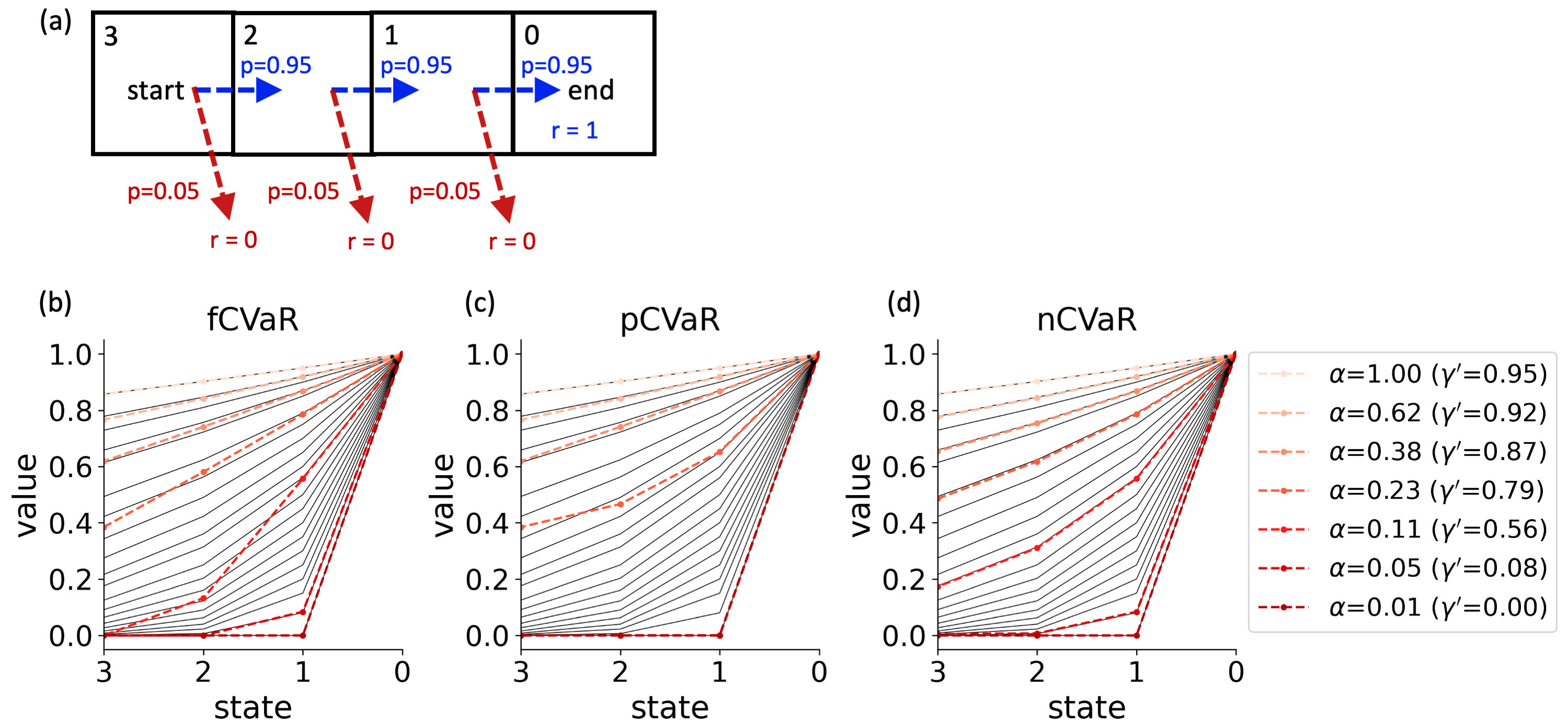}
  \caption{(a) A Markov evaluation problem with a single reward $r=1$ and a fixed $1-\gamma = 0.05$ probability of terminating in each step. (b-d) Increasing risk sensitivity (i.e., lower $\alpha$) leads to steeper discounting, but the shape of discounting differs for f-, p-, and nCVaR; f/pCVaR cut across the lines given by different values of $\gamma$, while nCVaR shows perfectly geometric discounting, with an implied discount factor that increases with decreasing $\alpha$.}
  \label{fig_3}
\end{figure}

\subsection*{Discussion}

Risk attitudes have subtle consequences in sequential problems, because probabilities can multiply over stages. We are only at the first stages of understanding how people's sensitivities operates in one-step problems, leaving much to be investigated about multi-step tasks. In particular, we need paradigms that can offer sharp differentiation.

Here, we adopted a coherent risk measure, CVaR, because of its disease-relevant focus on the most unfortunate potential outcomes \citep{gagne}, and showed how different sequentially-adapted versions, and different risk levels, might be distinguished behaviorally. Of course, subjects might not adopt exactly CVaR -- however, issues of additive and multiplicative compounding are likely to be critical to the analysis of any risk measure; and navigation tasks are likely to offer some of the most appealing experimental designs. We showed some stark, qualitative differences -- particularly between nCVaR and the other two measures.

Aside from administering these tasks to human subjects, recording choices and, potentially, neural activity, there are various compelling directions for investigation. One is the straightforward change to consider risk-seeking behavior (from the upper tail of the outcome distribution) rather than risk-averse behavior. A second is to consider more formally, and in a wider range of tasks, the relationship between dynamic risk preferences and hyperbolic discounting. A third is to consider ambiguity and unexpected uncertainty as well as risk. Finally, it would be important to consider psychopathologies, in different patient populations, which might partly be attributable to extreme values of $\alpha$.

\subsection*{Acknowledgements}

The authors have no competing interests to disclose. CG and PD are funded by the Max Planck Society. PD is also funded by the Alexander von Humboldt Foundation.

\bibliography{references}

\end{document}